\documentclass[journal, final]{IEEEtran}

\usepackage{epsfig}
\usepackage{graphics} 
\usepackage{amssymb}  
\usepackage{color}
\usepackage{tabularx}
\usepackage{lipsum}
\usepackage{psfrag}
\usepackage{placeins}
\usepackage{units}
\usepackage{subcaption}
\usepackage{soul}
\usepackage{bm}
\usepackage{pgfplots}
\usetikzlibrary{shapes,arrows,fit,calc}
\usepackage{amsmath}
\usepackage{multicol}
\usepackage{enumitem}
\usepackage{booktabs}
\usepackage{algorithm}
\usepackage[]{algpseudocode}

\tikzset{box/.style={draw, rectangle, rounded corners, thick,
node distance=12em, text width=6em, text centered,
minimum height=3.5em}}
\tikzset{cond/.style={draw, diamond, rounded corners, thick,
node distance=12em, text width=6em, text centered,
minimum height=2.5em}}
\tikzset{switch/.style={draw, circle, minimum size=3em, node distance=12em}, }

\tikzset{container/.style={draw, rectangle, dashed, inner sep=2em}}
\tikzset{line/.style={draw, thick, -latex}}
\tikzset{line2/.style={draw, thick}}

\usepackage[]{units}

\definecolor{mycolor1}{rgb}{0.00000,0.44700,0.74100}%
\definecolor{mycolor2}{rgb}{0.85000,0.32500,0.09800}%
\definecolor{mycolor3}{rgb}{0.92900,0.69400,0.12500}

\newcommand{\dt}{\delta t}

\usepackage{xcolor}

\newcommand{\iec}{\textit{i.e.},~}

\newcommand{\egc}{\textit{e.g.},~}

\newcommand{\eal}{\textit{et al.}~}

\newcommand{\fig}{Figure~}

\newcommand{\tabb}{Table~}

\newcommand{\sect}{Section~}

\usepackage{hyperref}
\usepackage{adjustbox}

\begin{document}
\title{Upper Body Pose Estimation Using Wearable Inertial Sensors and Multiplicative Kalman Filter}
\author{Tommaso Lisini Baldi$^{1}$,~
		Francesco Farina$^{2}$,~
		Andrea Garulli$^{1}$,~
		Antonio Giannitrapani$^{1}$,~
 		Domenico Prattichizzo$^{1,3}$% <-this % stops a space
\thanks{\textcopyright 2019 IEEE.  Personal use of this material is permitted.  Permission from IEEE must be obtained for all other uses, in any current or future media, including reprinting/republishing this material for advertising or promotional purposes, creating new collective works, for resale or redistribution to servers or lists, or reuse of any copyrighted component of this work in other works.}
\thanks{The research leading to these results has received funding
from the European Union's Horizon 2020 research and innovation programme under grant agreement n. 688857 of the project `` SOFTPRO - Synergy-based Open-source Foundations and Technologies for Prosthetics and RehabilitatiOn''
.}% 
\thanks{$^1$Tommaso Lisini Baldi, Andrea Garulli, Antonio Giannitrapani, and Domenico Prattichizzo are with the Department of Information Engineering and Mathematics, University of Siena, Via Roma 56, I-53100 Siena, Italy. $\{$lisini, garulli, giannitrapani, prattichizzo$\}$@diism.unisi.it}%
\thanks{$^2$Francesco Farina is with the Department of Electrical, Electronic and Information Engineering ``G. Marconi'', Universit{\`a} di Bologna, Bologna, Italy; franc.farina@unibo.it}
\thanks{$^3$Domenico Prattichizzo is with the Department of Advanced Robotics, Istituto Italiano di Tecnologia, Genova, 16163, Italy. {domenico.prattichizzo@iit.it}}
}

\maketitle

\begin{abstract}
Estimating the limbs pose in a wearable way may benefit multiple areas such as rehabilitation, teleoperation, human-robot interaction, gaming, and many more. Several solutions are commercially available, but they are usually expensive or not wearable/portable.
We present a wearable pose estimation system (WePosE), based on inertial measurements units (IMUs), for motion analysis and body tracking. Differently from camera-based approaches, the proposed system does not suffer from occlusion problems and lighting conditions, it is cost effective and it can be used in indoor and outdoor environments.
Moreover, since only accelerometers and gyroscopes are used to estimate the orientation, the system can be used also in the presence of iron and magnetic disturbances.
An experimental validation using a high precision optical tracker has been performed. Results confirmed the effectiveness of the proposed approach. 
\end{abstract}

\section{Introduction}

Wearable sensors have initially been employed as diagnostic and monitoring tools for gait analysis %
and joint kinematics. 
Nowadays, their main applications are still in the healthcare field~\cite{avci2010activity}, but new potential applications are emerging in: rehabilitation~\cite{mohammadi2016fingertip}, gaming~\cite{bleiweiss2010enhanced}, human robot interaction~\cite{baldi2017design}, human computer interface~\cite{meli2017hand}, human monitoring~\cite{mukhopadhyay2015wearable}, and many more.
In these applications, wearability represents a key feature because it does not constraint the user's motion and consequently improves the way users interact with each other and with the surrounding environment. In fact, wearable sensors have the advantage of being portable, lightweight and well integrable with other devices. Thanks to this, there is a growing interest in studying and developing novel wearable solutions to accurately track the human body. Unfortunately, most of the existing solutions are neither wearable nor portable since they usually rely on grounded/bulky hardware and/or structured environments. 

Optical tracking systems such as Vicon (Vicon Motion Systems, UK) and Optitrack (NaturalPoint Inc., USA) exploit active or passive optical markers to estimate the human body configuration with high precision and accuracy. The main drawback of these systems is the need of a structured environment. As another example, exoskeletons allow to accurately estimate the human pose thanks to their rigid structure and high quality sensors. Disadvantages usually concern cost and weight.

The aforementioned solutions provide very accurate motion estimation, but they usually have a high cost and they are neither wearable/portable nor usable in unstructured or outdoor scenarios. In order to favor wearability and reduce costs, camera-based tracking algorithms have become a widespread solution, thanks to the improvements in computer vision techniques and computational power. In~\cite{michel2017markerless}, the authors present a body tracker using commercial \mbox{RGB-D} cameras.
On the other hand, camera-based solution have some limitations as well: \mbox{RGB-D} cameras might not work properly in outdoor environments due to infrared interference and occasional occlusions may induce a poor estimation of the body posture.

\begin{figure}[t]
\centering
\null
\hfill
\begin{subfigure}{0.45\columnwidth}
\centering
\includegraphics[height=1.1\columnwidth, width=\columnwidth]{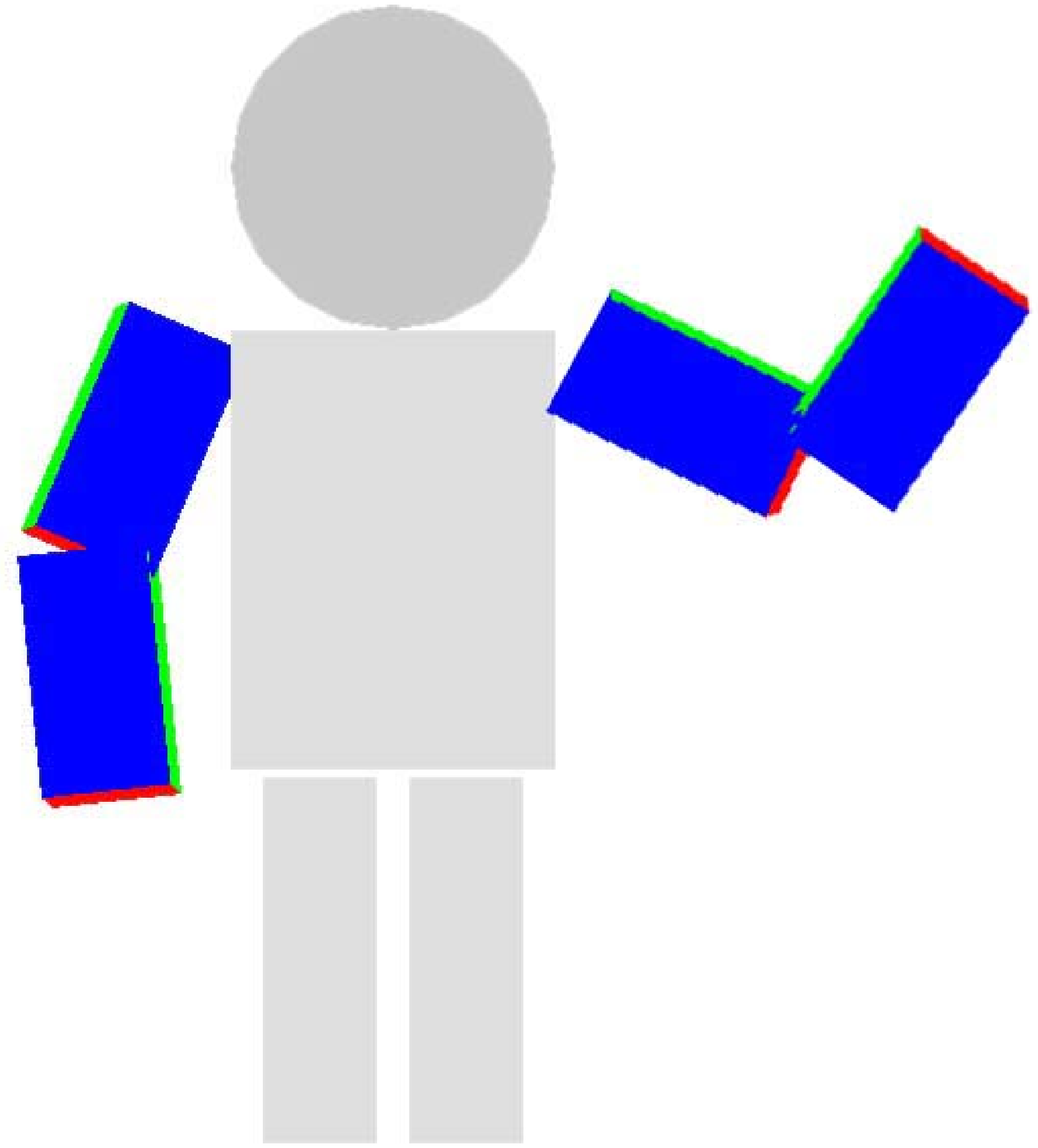}\label{FIG:1}
\end{subfigure}
\hfill
\begin{subfigure}{0.35\columnwidth}
\centering
\includegraphics[width=\columnwidth]{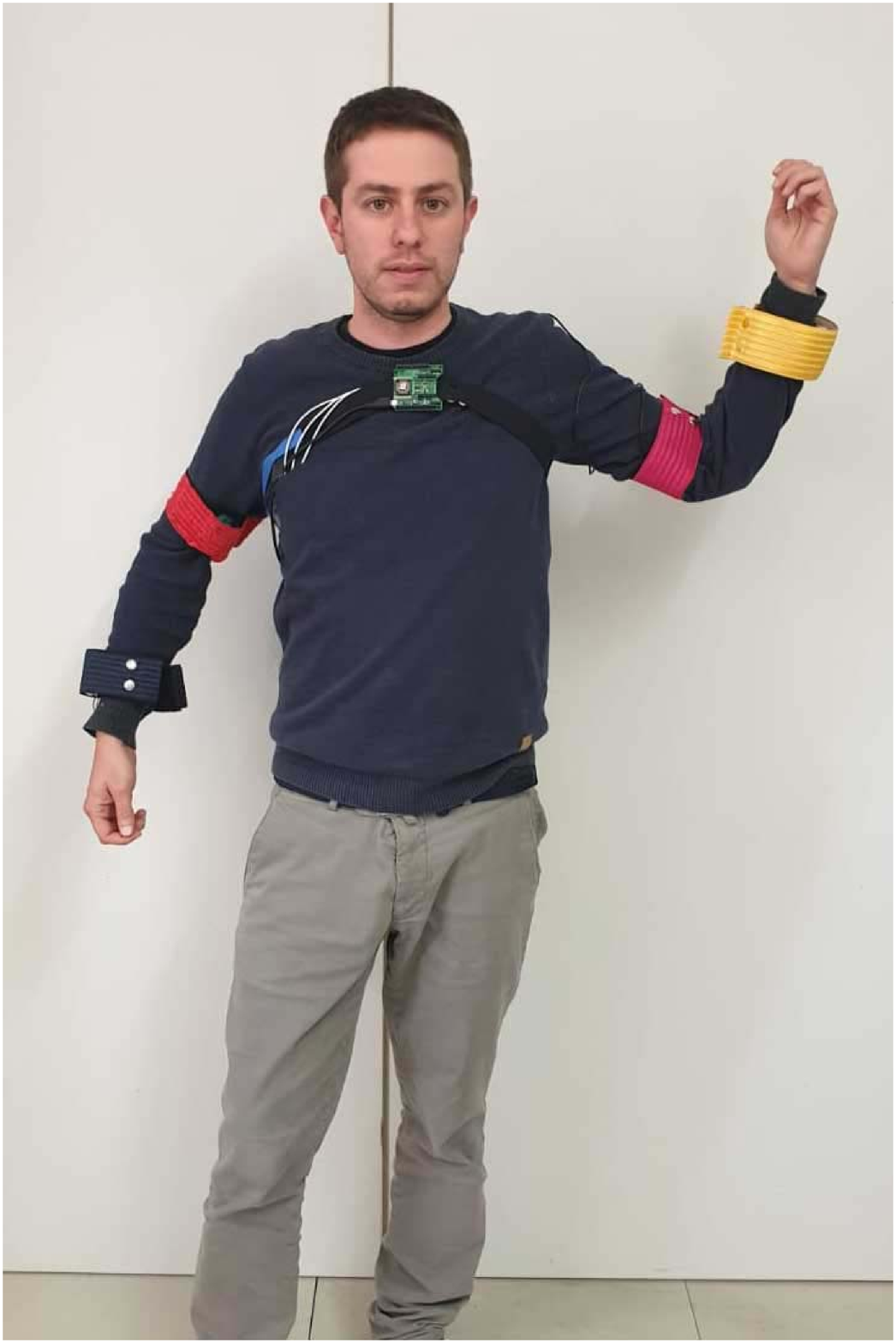}\label{FIG:2}
\end{subfigure}
\hfill\qquad
\null
\caption{Example of upper limbs pose reconstruction using the proposed wearable system.
}\label{FIG:concept}
\end{figure}

A viable solution to overcome these limitations consists in using fabric-integrated devices with rigid or flexible goniometers which are worn by the user~\cite{legnani2000model}. Nguyen \eal developed a sensing system, exploiting optical linear encoders, to measure limb joints angular data in home-based rehabilitation~\cite{nguyen2011wearable}. This type of solution uses kinematic reconstruction to determine the body posture. The weak point is that attachments of the body-based linkages as well as the positioning of the goniometers could generate several problems.
Body soft tissues allow the linkages position, relative to the body, to change when motion occurs. Moreover, a perfect alignment of the sensors with the joints is difficult, especially for joints with multiple degrees of freedom (\egc wrist and shoulder).

A further way to estimate the pose of the human body consists in using Micro Electro-Mechanical Systems (MEMS), such as Magnetic, Angular Rate, and Gravity (MARG) sensors, including triaxial gyroscopes, accelerometers, and magnetometers \cite{poitras2019validity, kieliba2018comparison}. These sensors can be easily integrated with wearable devices and can be used to reconstruct the pose of the human body through specific algorithms~\cite{morris1973accelerometry, foxlin1996inertial,el2015human}. 
Tracking systems based on this technology are commercially available and allow to track the whole body, in outdoor and indoor environments, under different lighting conditions and free from grounded hardware~\cite{roetenberg2009xsens,giansanti2005development}. However, a problem arises when integrating angular rates measured by MARG sensors. A tiny bias in the gyroscope output generates a huge drift in the orientation estimate~\cite{luinge2005measuring}.
For drift-free body orientation estimation, several methods have been developed combining the signals from inertial and magnetic sensors~\cite{7983378, zhao2012motion}. In the mentioned works, the accelerometer measurements are used to determine the direction of the local vertical by sensing the gravity acceleration, whereas magnetic measurements provide plane heading using the direction of the earth magnetic field. 
This approach has drawbacks in indoor environments, where the magnetic field is often heavily disturbed.

This paper presents a further step towards the capability of reconstructing the human body posture using wearable sensors.
Our contribution consists in presenting a method based on Inertial Measurement Units (IMUs), which are low-cost electronic devices  integrating MEMS sensors on a single board. Measurements coming from different sensors are integrated via the Multiplicative Extended Kalman Filter (MEKF), demonstrated by Markley~\cite{markley2003attitude}, to accurately estimate the body posture.
The use of Kalman Filter for orientation estimation is not a novelty and it has already been employed within several different settings. For instance in \cite{sabatelli2013double}, an angular estimation system that works with inertial measurement units is presented.
Additional examples can be found in \cite{won2009fastening}, \cite{li2017automatic}, and \cite{roetenberg2007ambulatory}.
An MEKF-based algorithm which performs the correction step only when the measurements used for the correction are sufficiently reliable, has been proposed in~\cite{sabatini2006quaternion}. Unlike the aforementioned approaches, in this work we propose a specialized version of this algorithm in which only measurements coming from the gyroscope and the accelerometer (and not from the magnetometer) are used. One of the main novelties is to use attitude estimation for upper limbs pose reconstruction, by exploiting kinematic models of human body.
To the best of our knowledge, the presented approach represents the first implementation of a MEKF for upper body tracking using only inertial measurements.

The rest of the paper is organized as follows. \sect\ref{SEC:algorithm} describes the proposed orientation estimation algorithm for a single rigid body. \sect\ref{SEC:experimental} reports the results of the experimental validation carried out to assess the accuracy of the estimation algorithm. \sect\ref{SEC:tracking} shows an application of the filter to upper body tracking. In \sect\ref{SEC:conclusion} some conclusions are drawn.

\section{Single body attitude estimation}
\label{SEC:algorithm}
In this section, a MEKF-based algorithm for estimating the orientation of a single rigid body is presented. This is instrumental to the multi-body tracking method proposed in \sect\ref{SEC:tracking}.
A possible approach in designing quaternion-based tracking algorithms is to update the quaternion by \emph{adding} to the current estimate a term depending on the angular rate weighted by the sampling time. By doing so, the obtained quaternion may have non-unitary norm, so that a normalization is needed to represent a pure rotation, thus introducing additional errors in the estimation process.
Conversely, the MEKF~\cite{markley2003attitude} exploits \emph{multiplications}, so that unitary norm is preserved by construction. 
In this work we follow the aforementioned approach. 
As a standard Kalman Filter, the MEKF consists of two main steps: a prediction step and a correction step. These steps are performed on a state vector containing the orientation error $a \in\Re^3 $ between the estimated attitude and the true one, and the gyroscope bias $b_\omega \in\Re^3$.
In the prediction step, the quaternion is updated along with the state vector and the covariance matrix. During the correction step, the state vector and the covariance matrix are updated by using the available measurements, while the quaternion estimate is corrected according to the new state vector.

The key point during the correction step is the selection of the measurements to use. 
In the considered setting, an IMU is attached to a moving rigid body.
The measurements coming from the magnetometer are not exploitable due to soft and hard iron disturbances, which are difficult to filter out (even by mapping the environment).  In manipulation tasks, for instance, it is impossible to predict, estimate, and remove distortion of the magnetic field induced by objects.
Moreover, during the body motion the measured acceleration can differ significantly from gravity acceleration, thus making the readings unusable for orientation estimation.

Here we assume the acceleration \textit{g} to be normalized with respect to the gravity acceleration. Thus, we say that if $||g||=1$ the accelerometer is measuring the gravity exactly. Hence, when $||g||\approx 1$ the measurements provided by the accelerometer can be used for orientation estimation. Such a condition is typically satisfied when the IMU is not moving.
Regarding the gyroscope readings, they are usually very accurate, despite of the body motion.

In the MEKF, angular rates are used for updating the attitude quaternion, during the prediction step. Moreover, gyroscope measurements can be employed for correcting the bias. In fact, when the IMU is almost steady, the gyroscope can be used to accurately measure its own bias $b_\omega$.

From the above discussion it is clear that measurements informative enough to be used in the correction step are, in general, not available when the body is moving. The proposed algorithm is a MEKF in which the correction step is performed only when informative measurements are available, \iec during time intervals in which the IMU is steady. We call these time intervals \emph{static phases} and we refer to the proposed algorithm as sMEKF.
In the following, we briefly outline the model and MEKF equation; the interested reader is referred to \cite{markley2003attitude} for details. 
We denote by $q$ the quaternion\footnote{We represent a quaternion as a 4 component vector $q=[q_w\, q_x\, q_y\, q_z]^\top$ where $q_w$ is the scalar part.} representing the attitude of a rigid body and by $\omega$ the angular rate of the body. 
The evolution of the attitude of the body can be expressed as
\begin{equation*}
\dot{q} =\frac{1}{2}q\otimes q(\omega) ,
\end{equation*}
where $\otimes$ denotes the quaternion multiplication operator and given a vector $v=[v_x\,v_y\,v_z]^\top$, we indicate by \mbox{$q(v)=[0\,v_x\,v_y\,v_z]^\top$} its \emph{quaternion form}.
Let $q_t$ and $\omega_t$ be the quaternion and the angular rate at time $t$, and $\dt$ be the sampling time. Under the assumption of small angles, the resulting attitude quaternion at time $t+\delta t$ is \begin{equation*}\label{eq:quat_prop}
	q_{t+\dt}=q_t\otimes q(\omega_t\dt).
\end{equation*}
The true attitude at time $t$ can be expressed as
\begin{equation*}\label{eq:qt}
 q_t=\hat{q}_t\otimes\delta q({a}_t),
\end{equation*}
where $\hat{q}_t$ is the current estimate of the quaternion and $\delta q(a_t)$ represents the rotation from $\hat{q}_t$ to the true attitude $q_t$, parametrized by the small angle vector $a_t$. 

We define the gyroscope and the accelerometer output, respectively, as $\omega_{out}$ and $g_{out}$ which we assume to be modeled as
\begin{equation}
\label{eq:measurements}
\begin{aligned}
	\omega_{out} &= \omega+ b_{\omega}+ w_{\omega},\quad \dot{b}_\omega=w_{b}\\
	g_{out} &= g + w_{g}
\end{aligned}
\end{equation}
where $\omega$ and $g$ are the true angular rate and acceleration, respectively, and $w_{\omega}$, $w_{g}$ and $w_b$ are disturbances modeled as white noises with zero mean and covariance matrices $\Sigma_{\omega}$, $\Sigma_{g}$ and $\Sigma_b$, respectively.

The aim of the MEKF is to estimate the 6-component state vector

\begin{equation*}
  x_t=
  \begin{bmatrix}
  a_t \\
  b_{{\omega},t}
  \end{bmatrix}
\end{equation*}
at each time $t$.
The estimated angular rate is defined as
	\begin{equation*}
		\hat{{\omega}}\triangleq{\omega}_{out}-\hat{{b}}_{\omega}
	\end{equation*}
being $\hat{{b}}_{\omega}$ the estimated bias. 

By following the derivation in \cite{markley2003attitude}, the state dynamics can be written as
	
\begin{equation}\label{xevol2}
\hspace{-1ex}{x}_{t+\delta t}\!=\!f(x_t,\!t)\!=\!\begin{bmatrix}\!{a}_t\!+\!\delta t\!\left(\!-\![\hat{{\omega}}_t]_\times{a}_t\!+\!\hat{{b}}_{\omega,t}\!-\!b_{\omega,t}\!-\!{w}_{\omega,t}\right)\!\\
    {b}_{\omega,t}+{w}_{b_{\omega,t}}\delta t
  \end{bmatrix},
\end{equation}	
being $[\bullet]_\times$ a skew-symmetric matrix used to represent cross products as matrix multiplications.

Defining $
y_{t+\delta t} = [{g}_{out_{t+\delta t}},\omega_{out_{t+\delta t}}]^\top
$, the measurement model \eqref{eq:measurements} gives
\begin{equation}\label{y2}
y_{t+\delta t} = h(x_{t+\delta t}) = 
\begin{bmatrix}
  {A}(q_{t+\delta t})^\top{g}_{I} + {w}_{g_{t+\delta t}}\\
  {\omega}_{t+\delta t}+{b}_{\omega_{t+\delta t}} + {w}_{\omega_{t+\delta t}}
\end{bmatrix}
\end{equation}
where ${g}_I=[0\,0\,1]^\top$ is the gravity acceleration in the inertial frame and $A(q)$ denotes the rotation matrix corresponding to the quaternion $q$.

\makeatletter
\renewcommand{\ALG@name}{Algorithm}
\makeatother

\algnewcommand\algorithmicinput{\textbf{Initialization:}}
\algnewcommand\init{\item[\algorithmicinput]}

\algnewcommand\algorithmicevol{\textbf{Evolution:}}
\algnewcommand\evol{\item[\algorithmicevol]}

\begin{algorithm}[t]
	\caption{sMEKF algorithm.}\label{TAB:mekf}
	\begin{algorithmic}{}
	\init $\hat{x}_{0\mid 0}$, $\hat{q}_{0\mid 0}$, $P_{0\mid 0}$, 
	${R}=
			\begin{bmatrix}
			{\Sigma}_g &{0}_{3\times 3}\\
			{0}_{3\times 3} &{\Sigma}_\omega
			\end{bmatrix}$
	
	\vspace{2ex}
	
	\evol
	
	\vspace{1ex}
	
	\For{all $t$} 
		
		\vspace{1ex}
		
	%	\State \textsf{{PREDICTION}}
	\State $\hat{\omega}_{t+\delta t | t} = \omega_{out,t} - \hat{b}_{\omega_{t|t}}$
		\vspace{1ex}
		
		\State $\hat{q}_{t+\delta t|t}= \hat{q}_{t|t}\otimes q(\hat{{\omega}}_{t+\delta t|t}\delta t)$ 
		
		\vspace{1ex}
		
		\State $\hat{{x}}_{t+\delta t | t}=
 					 \begin{bmatrix}
    									{0} \\
    								\hat{{b}}_{\omega_{t|t}}
  					\end{bmatrix}$
		
		\vspace{1ex}
					
		\State ${F}_t={I}_{6\times 6}+ \delta t
                          \begin{bmatrix}
                            -[\hat{{\omega}}_{t|t}]_\times & -{I}_{3\times 3} \\
                             {0}_{3\times 3} & {0}_{3\times 3}
                          \end{bmatrix}$
                   
		\vspace{1ex}
		       
                 \State  ${G}_t=\delta t 
                 	\begin{bmatrix}
  			  -{I}_{3\times 3} & {0}_{3\times 3} \\
  				  {0}_{3\times 3} & {I}_{3\times 3}
			  \end{bmatrix}$

		\vspace{1ex}
					
		\State ${P}_{t+\delta t|t}={F}_t{P}_{t|t}{F}_t^\top+{G}_t{Q} {G}_t^\top$

		\vspace{2ex}		

		 \If{Eq. \eqref{eq:cond1}-\eqref{eq:cond4} hold}
		 
		\vspace{1ex}

  		%\State \textsf{{CORRECTION}}
  
		\vspace{0.5ex}
		
                  \State ${H}_{t+\delta t}=
                		\begin{bmatrix}
               		 [{A}(\hat{q}_{t+\delta t|t})^\top{g}_I]_\times & {0}_{3\times 3} \\
               		 {0}_{3\times 3} & {I}_{3\times 3}
                		\end{bmatrix}$
		
		\vspace{1ex}

		\State ${K_{t+\delta t}}={P}_{t+\delta t|t}{H}_{t+\delta t}^\top[{H}_{t+\delta t}{P}_{t+\delta t|t}{H}_{t+\delta t}^\top+{R}]^{-1}$
		
		\vspace{1ex}

  		\State $\hat{{x}}_{t+\delta t|t+\delta t}=\hat{{x}}_{t+\delta t|t}+{K}_{t+\delta t}({y}_{t+\delta t}-h(\hat{{x}}_{t+\delta t|t}))$
  
		\vspace{1ex}
		
  		\State ${P}_{t+\delta t|t+\delta t}={P}_{t+\delta t|t}[{I}_{6\times 6}-{H}_{t+\delta t}^\top {K}_{t+\delta t}^\top]$
  
		\vspace{1ex}
		
   		\State  $\hat{q}_{t+\delta t|t+\delta t}=\hat{q}_{t+\delta t|t}\otimes\delta q(\hat{{a}}_{t+\delta t|t+\delta t}).$

 		 \Else
 		 \State $\hat{{x}}_{t+\delta t|t+\delta t} = \hat{{x}}_{t+\delta t|t}$
  
		\vspace{1ex}
		
 		 \State ${P}_{t+\delta t|t+\delta t}={P}_{t+\delta t|t}$
  
		\vspace{1ex}
		
		  \State  $\hat{q}_{t+\delta t|t+\delta t}=\hat{q}_{t+\delta t|t}$
  
		 \EndIf
 
 		\EndFor
  	
	\end{algorithmic}
\end{algorithm}

The sMEKF algorithm is reported in Algorithm \ref{TAB:mekf}, where the matrices $F_t$, $G_t$, and $H_{t+\delta t}$ are the Jacobian of the functions $f$ and $h$ in \eqref{xevol2} and \eqref{y2} (see \cite{markley2003attitude} for additional details).

In order to establish whether a correction step can be performed, we define a moving time window that is used to detect if the IMU is steady. The window size is denoted by
\begin{equation*}
  W=N\delta t,\quad N\in\mathbb{N}
\end{equation*}
where $N$ represents the number of samples in the time window. 
Given a certain time instant $\bar{t}$, we want to check if the IMU has been still for $t\in[\bar{t}-W,\bar{t}]$.
In order to do so, we verify that the following conditions are satisfied:
\begin{align}
S(\omega_{out,{\bar{t}}};W) \leq\alpha \hat{\Sigma}_\omega\label{eq:cond1} \\
S(g_{out,{\bar{t}}};W)\leq\beta \hat{\Sigma}_g\label{eq:cond2}\\
\left|E(\|{g}_{out,{\bar{t}}}\|; W)-1\right|\leq \gamma_{1}\label{eq:cond3}\\
S(\|{g}_{out,{\bar{t}}} \|;W) \leq \gamma_{2}\label{eq:cond4}
\end{align}
where, given a sequence $\{x_t\}$ with $x_t\in\Re^n$ for all $t$, we have denoted by $$E(x_{\bar{t}}; W)=\frac{1}{N}\sum_{t=\bar{t}-W}^{\bar{t}} x_t$$ the sample mean over a time window of length $W$, and by
%\begin{equation*}
\begin{align*}
S(x_{\bar{t}};W)=\frac{1}{N{-}1}\sum_{t=\bar{t}-W}^{\bar{t}}(x_t&-E(x_{\bar{t}}; W))(x_t-E(x_{\bar{t}}; W))^\top
\end{align*}
%\end{equation*}
the sample variance over the same time window.
The matrices $\hat{\Sigma}_\omega$ and $\hat{\Sigma}_g$ are the estimated covariance matrices of $w_\omega$ and $w_g$,  computed in an initial calibration phase in which the IMU is kept steady for a sufficient laying time and the parameters $\alpha,\beta,\gamma_{1},\gamma_{2}$ are fixed constants. 

Conditions~\eqref{eq:cond1} and~\eqref{eq:cond2} guarantee that the variances of the measurements are sufficiently close to the ones measured in the calibration phase (in which the IMU is steady). Condition~\eqref{eq:cond3} and~\eqref{eq:cond4} are used to check if the magnitude of the acceleration measured in the last $W$ steps has been sufficiently close to $1$.
Clearly, depending on the value of the constants $\alpha,\beta,\gamma_{1},\gamma_{2}$ and $\gamma_{3}$, conditions~\eqref{eq:cond1}-\eqref{eq:cond4} are more or less restrictive. The higher the chosen values, the more correction steps are performed. However, too high values increase the risk of detecting false static phases. In our experiments, we set $\alpha=\beta=2$ and $\gamma_1=\gamma_2=0.01$.

\begin{figure}[t]
\centering
\resizebox{0.9\linewidth}{!}{%\documentclass{standalone}
%
%\usepackage{tikz}
%\usetikzlibrary{shapes,arrows,fit,calc}
%
%\tikzset{box/.style={draw, rectangle, rounded corners, thick,
%node distance=12em, text width=6em, text centered,
%minimum height=3.5em}}
%\tikzset{cond/.style={draw, diamond, rounded corners, thick,
%node distance=12em, text width=6em, text centered,
%minimum height=2.5em}}
%\tikzset{switch/.style={draw, circle, minimum size=3em, node distance=12em}, }
%
%\tikzset{container/.style={draw, rectangle, dashed, inner sep=2em}}
%\tikzset{line/.style={draw, thick, -latex}}
%\tikzset{line2/.style={draw, thick}}
%
%
%\begin{document} 
\begin{tikzpicture}[auto]
\node [box] (gyro) {\huge $\omega_{out}$};
\node [box, below of=gyro] (acc) {\huge $g_{out}$};
\node[container , fit=(gyro) (acc)] (IMU) {};
\node at (IMU.north) [above,node distance=0 and 0] {\huge IMU};

\node [box, right of=gyro ] (propagation) {\LARGE Predict};
\path [line] (gyro) |- (propagation);

\node[coordinate, right of=propagation, xshift=4em] (decide) {};
\node[box, right of=propagation, xshift=8em, yshift=3em] (correction) {\LARGE Correct};
\node[right of=propagation, xshift=21em, yshift=-3em] (no_correction) {};

\path [line2] (propagation) -- (decide);

\node[right of=decide,yshift=3em] (decide1) {};
\node[right of=decide,yshift=-3em] (decide2) {};
\path [line2] (decide) -- (decide1.center);
\path [line2] (decide) -- (decide2.center);
\path [line] (decide1.center) -- (correction);
\path [line2] (decide2.center) -- (no_correction.center);

\node[right of=correction, xshift=6em] (output1) {};
\node[right of=no_correction, xshift=2em] (output2) {};
\node[switch, right of=correction, yshift=-3em] (decide2) {};
\path [line2] (correction) --(output1.center);
\path [line2] (no_correction.center) --(output2.center);
\path [line2] (output1.center) -- (decide2);
\path [line2] (output2.center) -- (decide2);

\node [cond, right of=acc, xshift=24em] (conditions) {\huge \eqref{eq:cond1}-\eqref{eq:cond4}};
\path [line] (IMU) --(IMU-|propagation) |- (conditions);
%\path [line] (conditions.north) |- (IMU -| decide.south) -- (decide.south);
\path [line] (conditions) -| (decide2.south);

\node[container , fit=(propagation) (decide) (correction) (decide2) (no_correction) (conditions)] (sMEKF) {};
\node at (sMEKF.north) [above,node distance=0 and 0] {\huge sMEKF};

\node[right of=decide2, xshift=6em] (output) {\huge $\hat{q}_{t+\delta t|t+\delta t}$};
\path [line] (decide2) -- (output);
\end{tikzpicture}
%\end{document}
}
\caption{Overview of the proposed tracking system.}\label{FIG:flowchart}
\end{figure}
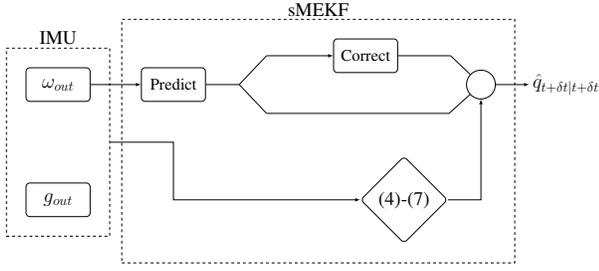

If at a given time, conditions~\eqref{eq:cond1}-\eqref{eq:cond4} hold, a correction step of the sMEKF (as detailed in Algorithm~\ref{TAB:mekf}) is performed. Otherwise, no correction occurs and the filter proceeds to the next prediction step. 
A flow chart representing the proposed algorithm is reported in Figure~\ref{FIG:flowchart}.

\begin{figure}[t]
	\centering
	\small
	\resizebox{0.95\linewidth}{!}{
	\input{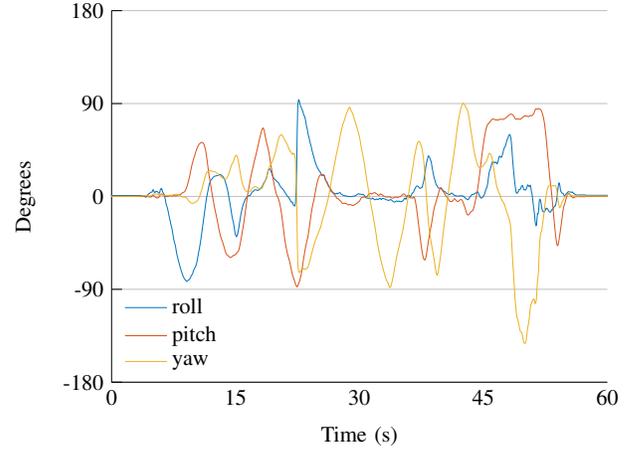}
	}
	\caption{Dynamic validation. The MARG sensor was positioned onto a flat platform
	together with seven passive optical markers. We kept the platform steady for \unit{5}{s},
	then we freely moved and rotated it for \unit{50}{s}, and finally, we kept it steady for
	further \unit{5}{s}. A representative trial is depicted; only the orientation estimated by
	the optical tracker is reported.
	\label{FIG:representativeQ}}
\end{figure}

\section{Experimental Validation}
\label{SEC:experimental}

\begin{figure*}[t]
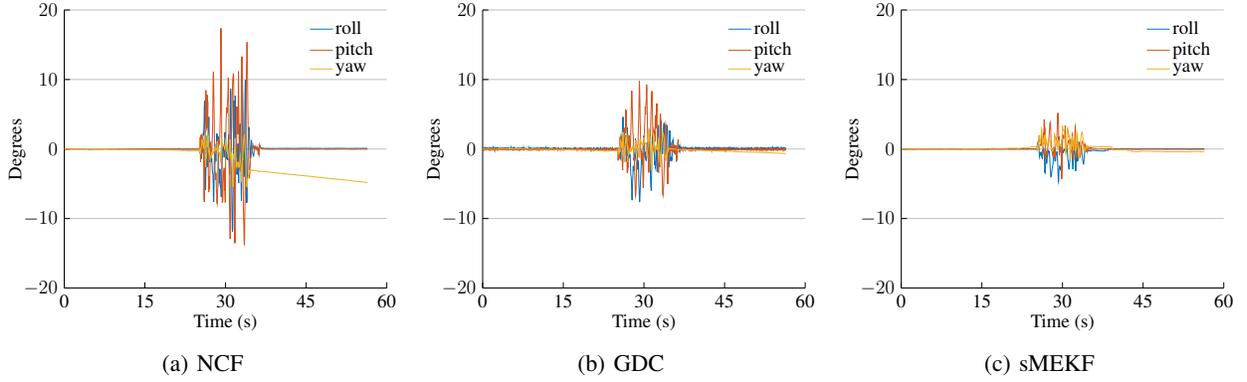

\centering
\Large
\begin{subfigure}{0.3\textwidth}
\resizebox{\linewidth}{!}{
 \input{pics/mahoney_drift}
 }
 \caption{NCF}
\label{FIG:Drift_NCF}
\end{subfigure}
\begin{subfigure}{0.3\textwidth}
\resizebox{\linewidth}{!}{
 \input{pics/madgwick_drift}
 }
 \caption{GDC}
 \label{FIG:Drift_GDC}
\end{subfigure}
\begin{subfigure}{0.3\textwidth}
\resizebox{\linewidth}{!}{
 \input{pics/mekf_drift}
 }
 \caption{sMEKF}
 \label{FIG:Drift_EKF}
\end{subfigure}
       \caption{Drift validation. Roll, Pitch, Yaw angle errors for a representative trial. \label{FIG:Drift_Errors}}
\end{figure*} 

\begin{figure*}[t]
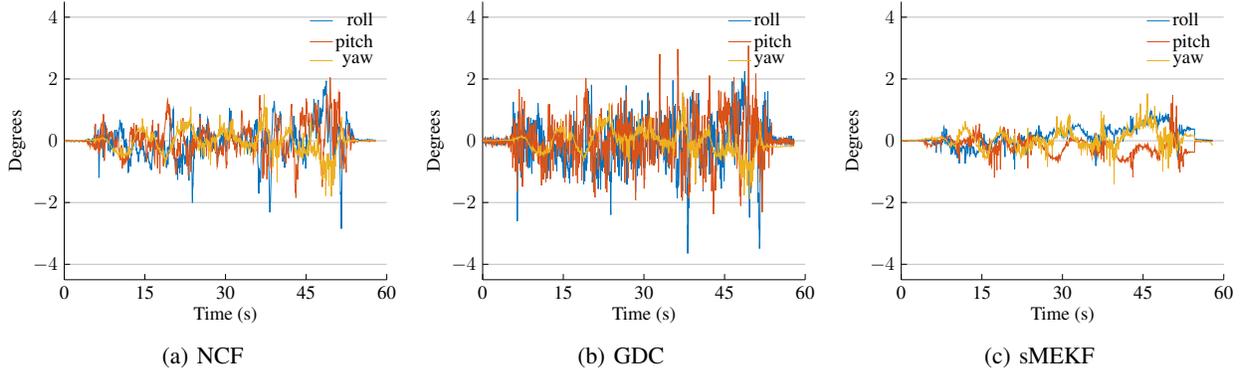

\centering
\Large
\begin{subfigure}{0.3\textwidth}
\resizebox{\linewidth}{!}{
 \input{pics/mahoney}
 }
 \caption{NCF}
 \label{FIG:ErrorNCF}
\end{subfigure}
\begin{subfigure}{0.3\textwidth}
\resizebox{\linewidth}{!}{
 \input{pics/madgwick}
 }
 \caption{GDC}
 \label{FIG:ErrorGDC}
\end{subfigure}
\begin{subfigure}{0.3\textwidth}
\resizebox{\linewidth}{!}{
 \input{pics/mekf}
 }
 \caption{sMEKF}
\label{FIG:ErrorEKF}
\end{subfigure}
       \caption{Dynamic validation. Roll, Pitch, Yaw angle errors for a representative trial.\label{FIG:AlgorithmsError_DYN}}
\end{figure*}

In this section we report the experimental validation of the sMEKF algorithm for estimating the orientation of a single body.  
Performance of the proposed  method are compared to two widely used algorithms for this purpose \cite{amaro2016survey}:

\begin{enumerate}[label=(\roman*)]
\item the Nonlinear Complementary Filter (NCF), proposed by Mahoney~\cite{mahony2008nonlinear};
\item the Gradient Descent algorithm coupled with a Complementary filter (GDC), proposed by Madgwick~\cite{madgwick2011estimation}.
\end{enumerate}  

Two different experiments, evaluating the accuracy of the proposed method, have been carried out.
The first intends to show the lack of drift, whereas the second demonstrates the accuracy in tracking both slow and rapid body movements.
It is important to highlight that the performance of the considered approaches largely depends on the integration capability of the sensors, thus the higher is the sampling rate, the more accurate is the estimation. Two different sampling rates for each experiment have been considered:\textit{ i)} high sampling rate (\unit[1]{kHz}), and \textit{ii)} low sampling rate (\unit[100]{Hz}). A single IMU (Xsens \mbox{MTI-3}) has been placed on a flat platform together with seven passive reflective markers for the validation.

Tracking errors are computed with respect to a high accuracy optical tracking system, considered as the ground truth. A Vicon system, consisting of ten Bonita cameras and Tracker 3.7.0 software, has been used for this purpose.  
A preliminary calibration phase has been performed to align the reference frame of the IMU with the one of the Vicon. Then, the platform has been freely moved around, without any constraint. Raw data have been collected and post-processed using the three algorithms.
A single PC was in charge of collecting data streamed by the Vicon system and the inertial sensors. Ground truth data were gathered through the network using a high performance Ethernet card (HP Intel Ethernet I210-T1 GbE NIC), whereas the sensors transmitted raw data via a dedicated USB 3.0 port. Inertial raw data were then processed by the three algorithms and compared with the real orientation. This configuration minimized transmission delays, and packet misalignment.%, 

\subsection{Drift validation}

Firstly, experiments concerning drift compensation have been performed. Ten trials have been performed using the highest sampling rate (\unit[1]{kHz}). The platform carrying the IMU and the markers has been kept steady for 25 seconds, then it has been quickly moved, shacked, and rotated for \unit{10}{} seconds. Finally, it has been kept steady for further 25 seconds in order to quantify the error drift. It is worth pointing out that the motion of the platform has been over-stressed in this experiment. Thus, the error in estimating the orientation is affected also by the Vicon relatively slow sampling rate for all the algorithms.
In~\fig\ref{FIG:Drift_Errors} the results of a representative trial are reported. 
It can be seen that all the algorithms show negligible error drift as long as roll and pitch are concerned. However, the NCF algorithm presents a remarkable yaw drift. On the contrary, the GDC algorithm show a limited drift, while the one of sMEKF is negligible. 

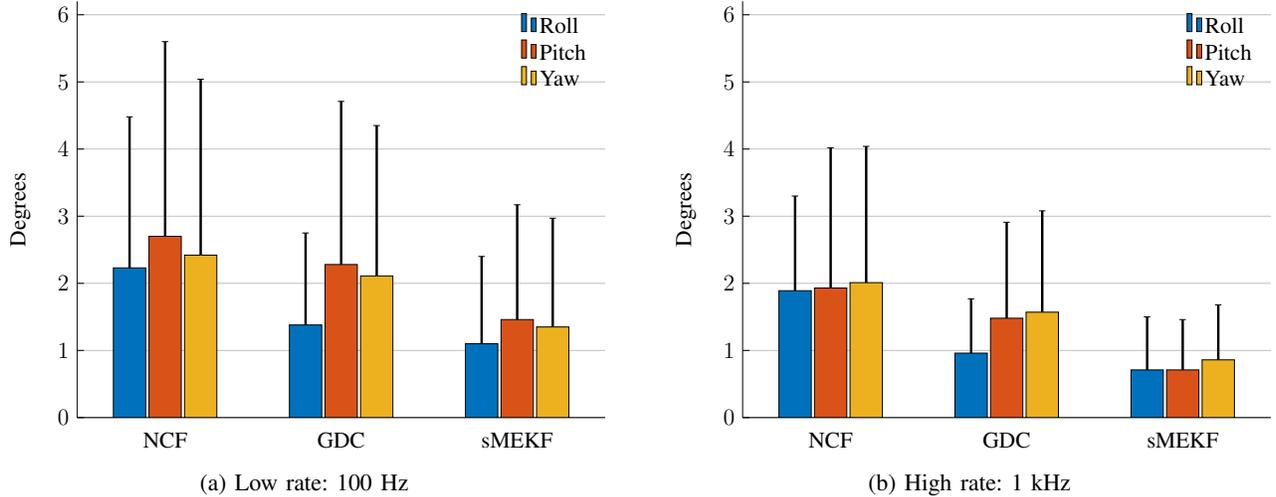
\begin{figure*}[t]
\begin{center}
\Large
\null\hfill
\begin{subfigure}{0.45\textwidth}
\resizebox{\linewidth}{!}{
 %\documentclass[border=10pt]{standalone}
%\usepackage{pgfplots}
%\pgfplotsset{width=7cm,compat=1.8}
%
%\definecolor{mycolor1}{rgb}{0.00000,0.44700,0.74100}%
%\definecolor{mycolor2}{rgb}{0.85000,0.32500,0.09800}%
%\definecolor{mycolor3}{rgb}{0.92900,0.69400,0.12500}
%\begin{document}
\begin{tikzpicture}

\begin{axis}[%
width=4.521in,
height=3.566in,
at={(0.758in,0.481in)},
scale only axis,
bar shift auto,
every outer x axis line/.append style={black},
every x tick label/.append style={font=\color{black}},
every x tick/.append style={black},
xmin=0.5,
xmax=3.5,
xtick={1,2,3},
xticklabels={{NCF},{GDC},{sMEKF}},
xlabel={},
every outer y axis line/.append style={black},
every y tick label/.append style={font=\color{black}},
every y tick/.append style={black},
ymin=0,
ymax=6.2,
ylabel={Degrees},
axis background/.style={fill=white},
axis x line*=bottom,
axis y line*=left,
legend style={legend cell align=left, align=left, fill=none, draw=none},
major x tick style = transparent,
ybar=2pt,
bar width=20pt,
ymajorgrids = true,
scaled y ticks = false,
]

\addplot[style={fill=mycolor1},error bars/.cd, y dir=plus, y explicit, error bar style={line width=1.5pt,solid}]
          coordinates {
          (1, 2.23) += (0,2.25) -= (0,2.25)
          (2, 1.38) += (0,1.37) -= (0,1.37)
          (3, 1.10) += (0,1.30) -= (0,1.30)
          };

\addplot[style={fill=mycolor2},error bars/.cd, y dir=plus, y explicit,error bar style={line width=1.5pt,solid},]
          coordinates {
          (1, 2.70) += (0,2.90) -= (0,2.90)
          (2, 2.28) += (0,2.43) -= (0,2.43)
          (3, 1.46) += (0,1.71) -= (0,1.71)
          };

\addplot[style={fill=mycolor3},error bars/.cd, y dir=plus, y explicit,error bar style={line width=1.5pt,solid},]
          coordinates {
          (1, 2.42) += (0,2.62) -= (0,2.62)
          (2, 2.11) += (0,2.24) -= (0,2.24)
          (3, 1.35) += (0,1.62) -= (0,1.62)
          };

\legend{Roll, Pitch, Yaw}

%\addplot [color=black, line width=2.0pt, draw=none, forget plot]
% plot [error bars/.cd, y dir = both, y explicit]
% table[row sep=crcr, y error plus index=2, y error minus index=3]{%
%0.777777777777778	1.78696666666667	1.65276666666667	1.65276666666667\\
%1.77777777777778	0.8164	0.691066666666667	0.691066666666667\\
%2.77777777777778	0.3606	0.4782	0.4782\\
%};
%\addplot [color=black, line width=2.0pt, draw=none, forget plot]
% plot [error bars/.cd, y dir = both, y explicit]
% table[row sep=crcr, y error plus index=2, y error minus index=3]{%
%1	1.2313	1.54133333333333	1.54133333333333\\
%2	0.8738	1.01753333333333	1.01753333333333\\
%3	0.2964	0.447433333333333	0.447433333333333\\
%};
%\addplot [color=black, line width=2.0pt, draw=none, forget plot]
% plot [error bars/.cd, y dir = both, y explicit]
% table[row sep=crcr, y error plus index=2, y error minus index=3]{%
%1.22222222222222	1.162	1.44936666666667	1.44936666666667\\
%2.22222222222222	0.915833333333333	1.14343333333333	1.14343333333333\\
%3.22222222222222	0.3314	0.379333333333333	0.379333333333333\\
%};
%\addplot[ybar, bar width=0.222, fill=mycolor2, draw=mycolor2] table[row sep=crcr] {%
%1	1.162\\
%2	0.915833333333333\\
%3	0.3314\\
%};
%\addlegendentry{Roll}
%
%\addplot[ybar, bar width=0.222, fill=mycolor3, draw=mycolor3] table[row sep=crcr] {%
%1	1.2313\\
%2	0.8738\\
%3	0.2964\\
%};
%\addlegendentry{Pitch}
%
%\addplot[ybar, bar width=0.222, fill=mycolor1, draw=mycolor1] table[row sep=crcr] {%
%1	1.78696666666667\\
%2	0.8164\\
%3	0.3606\\
%};
%\addlegendentry{Yaw}
%
%\addplot [color=black, forget plot]
%  table[row sep=crcr]{%
%0.5	0\\
%3.5	0\\
%};
\end{axis}
\end{tikzpicture}%

%\end{document}
 }
 \caption{Low rate: 100 Hz}
\label{FIG:errors100}
\end{subfigure} \hfill
\begin{subfigure}{0.45\textwidth}
\resizebox{\linewidth}{!}{
 %\documentclass[border=10pt]{standalone}
%\usepackage{pgfplots}
%\pgfplotsset{width=7cm,compat=1.8}
%
%\definecolor{mycolor1}{rgb}{0.00000,0.44700,0.74100}%
%\definecolor{mycolor2}{rgb}{0.85000,0.32500,0.09800}%
%\definecolor{mycolor3}{rgb}{0.92900,0.69400,0.12500}
%\begin{document}
\begin{tikzpicture}

\begin{axis}[%
width=4.521in,
height=3.566in,
at={(0.758in,0.481in)},
scale only axis,
bar shift auto,
every outer x axis line/.append style={black},
every x tick label/.append style={font=\color{black}},
every x tick/.append style={black},
xmin=0.5,
xmax=3.5,
xtick={1,2,3},
xticklabels={{NCF},{GDC},{sMEKF}},
xlabel={},
every outer y axis line/.append style={black},
every y tick label/.append style={font=\color{black}},
every y tick/.append style={black},
ymin=0,
ymax=6.2,
ylabel={Degrees},
axis background/.style={fill=white},
axis x line*=bottom,
axis y line*=left,
legend style={legend cell align=left, align=left, fill=none, draw=none},
major x tick style = transparent,
ybar=2pt,
bar width=20pt,
ymajorgrids = true,
scaled y ticks = false,
]

%ROLL
\addplot[style={fill=mycolor1},error bars/.cd, y dir=plus, y explicit, error bar style={line width=1.5pt,solid}]
          coordinates {
          (1, 1.89) += (0,1.41) -= (0,1.41) %NCF
          (2, 0.96) += (0,0.81) -= (0,0.81) %GDC
          (3, 0.71) += (0,0.79) -= (0,0.79) %sMEKF
          };  

        %pitch  
\addplot[style={fill=mycolor2},error bars/.cd, y dir=plus, y explicit,error bar style={line width=1.5pt,solid},]
          coordinates {
          (1, 1.93) += (0,2.09) -= (0,2.09)
          (2, 1.48) += (0,1.43) -= (0,1.43)
          (3, 0.71) += (0,0.75) -= (0,0.75)
          };

%yaw
\addplot[style={fill=mycolor3},error bars/.cd, y dir=plus, y explicit,error bar style={line width=1.5pt,solid},]
          coordinates {

          (1, 2.01) += (0,2.03) -= (0,2.03)
          (2, 1.57) += (0,1.51) -= (0,1.51)
          (3, 0.86) += (0,0.82) -= (0,0.82)
          };

\legend{Roll, Pitch, Yaw}

\end{axis}
\end{tikzpicture}%
%\end{document}
 }
 \caption{High rate: 1 kHz}
 \label{FIG:errors1K}
\end{subfigure}\hfill\null
\caption{Comparison among the three estimation algorithms. 
For each algorithm, the error mean and standard deviation in orientation estimation are reported, expressed as roll, pitch and yaw angles.\label{FIG:AlgorithmsError}}
\end{center}
\end{figure*}

\subsection{Dynamic validation}

The second experimental campaign aimed at verifying the capability of the proposed approach to correctly estimate orientation in a dynamic situation. In these experiments the platform has been randomly moved and rotated in space, simulating common body links motions. Twelve trials lasting 60 seconds each were performed.
We kept the platform steady for \unit{5}{s}, then we freely moved and rotated it for \unit{50}{s} and, finally, we kept it steady for further \unit{5}{s}. 
A representative motion of the platform is depicted in \fig\ref{FIG:representativeQ}.
A comparative error analysis between the three algorithms is reported in \fig\ref{FIG:errors100}, 
 and \fig\ref{FIG:errors1K} 
for \unit[100]{Hz} and \unit[1]{kHz} sampling rate, respectively. The mean and the standard deviation of the estimation error for \unit[100]{Hz} and \unit[1]{kHz} are reported in \tabb\ref{TABLE:AlgorithmsError_LR} and \tabb\ref{TABLE:AlgorithmsError_HR}, respectively.
In \fig\ref{FIG:AlgorithmsError_DYN} the orientation estimation errors in Euler angles for a single experiment at \unit[1]{kHz} are reported.
As it can be seen, the sMEKF performs better than the NCF and the GDC algorithms at both acquisition rates. In particular, at \unit[1]{kHz}, the mean and standard deviation of the attitude estimation error are significantly lower with respect to the NCF and the GDC algorithms (up to 61\% and 42\%, respectively). The lower sampling rate gives less   percentage difference in orientation estimation error: 47\% with respect to NFC and 32\% compared to GDC.

\section{Upper Limbs Tracking Application}
\label{SEC:tracking}
The attitude estimation scheme previously described  has been employed within a wearable system for body tracking. 
Kinematic models and human joint angles estimation have been combined to continuously reconstruct the body posture. In the previous section we described how to estimate the attitude of a single rigid body. Assuming two consecutive links connected by a spherical joint, we can estimate the angle between the two links through the IMUs attitudes. In what follows, we investigate the performance of sMEKF for upper body tracking.
To determine the body and limbs configuration, we use a well known biomechanical modeling technique based on a sequence of links connected by joints. This type of model allows the representation of any part of the human body (or robotic arm). To obtain a systematic method for describing position and orientation of each pair of consecutive links, we generate a homogeneous transformation matrix between the two links, by using the Denavit and Hartenberg method, following the approach described in~\cite{el2011upper}.
The homogeneous matrix is obtained by combining the link length and the quaternion computed by the sMEKF algorithm.
If each couple of consecutive links is related via a matrix, then, using the kinematic chain rule, it is possible to connect any link to another one (\egc the arm and forearm). 
\begin{table}
\small
\begin{center}
\begin{tabular}{|c|c|c|c|}
\hline
Algorithm & Roll (deg.) & Pitch (deg.) & Yaw (deg.)\\ \hline
NCF & 2.23 $\pm$ 2.25 & 2.70 $\pm$ 2.90 & 2.42 $\pm$ 2.62\\ \hline
GDC & 1.38 $\pm$ 1.37 & 2.28 $\pm$ 2.43 & 2.11 $\pm$ 2.24\\ \hline
MEKF & 1.10 $\pm$ 1.30 & 1.46 $\pm$ 1.71 & 1.35 $\pm$ 1.62\\ \hline
\end{tabular}
\caption{Mean and standard deviation of the attitude estimation error for the three considered algorithms sampled at 100~Hz.	\label{TABLE:AlgorithmsError_LR}}
\end{center}
\end{table}

\begin{figure}[b]
\centering
\includegraphics[width=0.98\columnwidth]{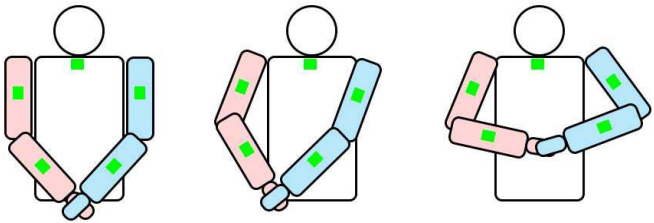}
\caption{Links length calibration: lengths are refined by solving the closed kinematic chain. Inertial sensors are indicated with green rectangles. \label{FIG:Calibration}}
\end{figure}

A common problem in wearable tracking systems employing sensors attached to human limbs is that the initial pose between the sensor and the body segment is unknown \cite{miezal2016inertial}. Moreover, the computation of limb trajectories by numerical integration of acceleration signals is not a reliable approach, because of both the unknown initial position and the noise affecting the measurements. This implies that standard tracking methods have large errors in correctly estimating the body posture. 
A main advantage of the sMEKF-based approach is that it does not require neither a joint-sensor calibration nor a joint approximation. Since each sensor estimates the actual orientation with respect to the initial pose, only the length of the links is required. Such lengths can be estimated in several ways in the calibration phase. 
Similarly to what is proposed in~\cite{roetenberg2009xsens}, we asked subjects to hold their hands together and freely move them (see \fig\ref{FIG:Calibration}). The distance between the left and the right palm is zero, so that one can take advantage of this closed kinematic chain to tune the model. Starting from an initial value (taken from anthropometric measurements~\cite{nasa1987man}), an optimization algorithm is used to refine the estimation of the links length. The \textit{a priori} lengths of the bones are used as a starting point to initialize the optimization procedures, which minimizes the distance between the two hands.

\begin{table}[b]
\normalsize
\begin{center}
\begin{tabular}{|c|c c|}
\hline
Algorithm & Mean (mm) & STD (mm) \\ \hline
NCF & 17.71 & 10.75 \\ \hline
GDC & 17.96 & 10.82 \\ \hline
MEKF & 12.05 & 6.20 \\ \hline
\end{tabular}
\caption{Mean and standard deviation of the hand pose reconstruction error.
	\label{TAB:traject_error}}
\end{center}
\end{table}

\begin{figure*}[t]
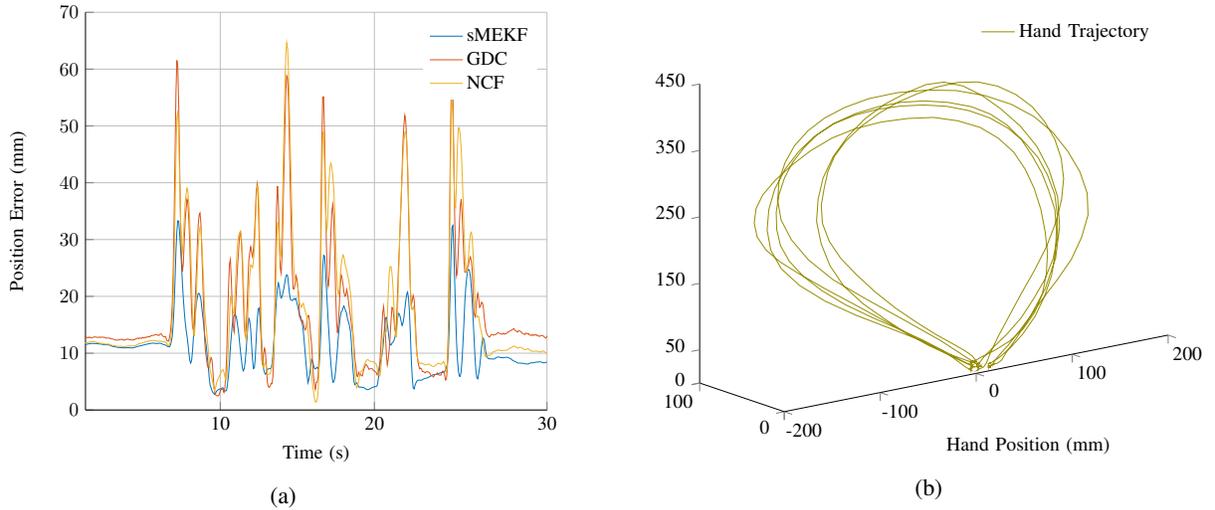

\centering
\null\hfill
\begin{subfigure}{0.42\textwidth}
\resizebox{\linewidth}{!}{
\input{pics/traject_error}
 }
 \caption{\label{FIG:traject_error}}
\end{subfigure}%
\hfill%
\begin{subfigure}{0.42\textwidth}
\resizebox{\linewidth}{!}{
 \input{pics/trajectory.tex}
 }
 \caption{\label{FIG:trajectory}}
\end{subfigure}%
\hfill%
\null
\caption{Hand pose reconstruction comparison among the three estimation algorithms. Panel (a) reports the hand tracking error in estimating the position with respect to the ground truth given by a Vicon system. In (b) a representative  trial is depicted.
\label{FIG:BodyTracking}}
\end{figure*}

In order to validate the proposed algorithm, we applied it to a real world scenario. The aim of the experiment is to track the motion of the hand of the subject by using the attitude of the IMUs and the links length. Differently from the calibration phase, this experiment is performed in open loop kinematic: neither closed loop nor additional constraints were exploited to reinforce the algorithms. Ten subjects took part in the experimental evaluation. Each subject was asked to place his hand on a table, keep it steady for a couple of seconds and then to draw six circles in the air and place the hand on the table at the end of each circle (representative hand motion is reported in \fig\ref{FIG:trajectory}).
Each participant wore five IMUs attached, respectively, to the chest, the arms and the forearms (as depicted in \fig\ref{FIG:Calibration}).
Once the skeleton dimensions had been estimated with the aforementioned procedure, the performance of each algorithm was evaluated by comparing the resulting trajectory with that obtained from a Vicon system used as ground truth.
Inertial raw data were collected and then post-processed using the three different algorithms. Each algorithm was used both to estimate the links length in the preparatory phase, and then in the following trials for reconstructing the body posture.
The system kinematic model is translated to body segment kinematics using a biomechanical model which assumes that the subject body includes body segments linked by joints and that the sensors are attached to the subject.

\begin{table}
\small
\begin{center}
\begin{tabular}{|c|c|c|c|}
\hline
Algorithm & Roll (deg.) & Pitch (deg.) & Yaw (deg.)\\ \hline
NCF & 1.89 $\pm$ 1.41 & 1.93 $\pm$ 2.09 & 2.05 $\pm$ 2.03\\ \hline
GDC & 0.96 $\pm$ 0.81 & 1.48 $\pm$ 1.43 & 1.57 $\pm$ 1.50\\ \hline
MEKF & 0.71 $\pm$ 0.79 & 0.73 $\pm$ 0.75 & 0.86 $\pm$ 0.82\\ \hline
\end{tabular}
\caption{Mean and standard deviation of the attitude estimation error for the three considered algorithms sampled at 1~kHz.
	\label{TABLE:AlgorithmsError_HR}}
\end{center}
\end{table} 

Joint origins are determined by the anatomical frame and are defined in the center of the functional axes with the directions of the X, Y and Z being related to functional movements.
We consider each joint as a spherical joint, enabling 3D motion for each segment. This strategy overcomes the problem of modeling troublesome joints like the distal radioulnar articulation. 
The independent orientation estimation of each body segment is a central benefit and it allows us to avoid the articulation to joint mapping. Such a mapping, in fact, is usually a great source of error and uncertainty due to the impossibility of a perfect modeling with a limited number of parameter.
The position error with respect to the ground truth of a representative trial is reported in \fig\ref{FIG:traject_error}.
It is worth pointing out that a part of the error can be attributed to link length estimates that are not perfect. In fact, if the link lengths are 5\% different from the real ones, an error in the attitude of at 1 degree (as it is the one of the sMEKF in \fig\ref{FIG:traject_error}) produces a maximum error in the trajectory of about 4 cm for a subject with an arm and a forearm of 30 cm each. 
The results of the  experimental evaluation confirm the expected performance. In \tabb\ref{TAB:traject_error} we report mean and standard deviation of the hand pose error with respect to the ground truth, computed among all the performed trials.

\section{Conclusion and future work}
\label{SEC:conclusion}
An innovative wearable and reliable tracking system enabling 3D motion capture in daily activities has been presented. The system utilizes inertial sensors to track the desired body portion in any environment, indoors or outdoors, allowing voluntary movements to be recorded and viewed on a standard PC in real-time. The average error in tracking is lower than \unit[1]{deg} sampling at \unit[1]{kHz}, and less than $1.50$ degree at \unit[100]{Hz}. Moreover, thanks to the system modularity, any body part can be tracked and reconstructed. The capability of the system goes further. It can be used to track any kinematic chain with known parameters. The absence of magnetic referenced measures allows the system to gather data also from metal objects or structure containing motors, like robotic arms and platforms. A \textit{\texttt{c++}} implementation of the algorithm is freely available with an open source license \cite{WearBoPosE}.

\bibliographystyle{IEEEtran}
\bibliography{biblio/IEEEfull,biblio/conference,biblio/biblio}

\end{document}